\documentclass[runningheads]{llncs}
\usepackage[T1]{fontenc}
\usepackage{amsmath,amssymb,amsfonts}
\usepackage{booktabs}
\usepackage{graphicx}
\usepackage{multirow}
\usepackage{xcolor}
\usepackage{algorithm}
\usepackage{algorithmic}
\usepackage{enumitem}
\usepackage{subcaption}
\usepackage{array}

\newcommand{\fb}{\textsc{FinBERT}}
\newcommand{\llm}{\textsc{LLaMA}}
\newcommand{\allxgb}{\textsc{All+XGB}}

\begin{document}

\title{Beyond Sentiment: Structured Information Extraction from Financial News}
\titlerunning{Beyond Sentiment: Structured Extraction from Financial News}

\author{
    Daohan Zhu\inst{1}
    \and
    Sitong Ge\inst{1}
    \and
    Ruofei Wang\inst{1}
    \and
    Honggu Chen\inst{1}
    \and
    Yubo Hou\inst{1}
    \and
    Tao Wan\inst{2}
    \and
    Zengchang Qin\inst{1,3}
}

\authorrunning{D. Zhu et al.}

\institute{
    School of ASEE, Beihang University, Beijing, China \\
    \and
    School of BME, Beihang University, Beijing, China \\
    \and
    CAIR and CECS, VinUniversity, Hanoi, Vietnam \\
    \email{zhudaohan@buaa.edu.cn} \quad
    \email{*zcqin@buaa.edu.cn}
}







\date{}

\maketitle

\begin{abstract}
Financial sentiment analysis has become a standard component in news-driven stock prediction, yet it reduces rich, multi-dimensional news articles to a single polarity score. We hypothesize that financial news encodes multiple orthogonal information dimensions---event type, impact scope, temporal horizon, and semantic confidence---that sentiment alone cannot capture, and that these dimensions carry independent predictive value. To test this hypothesis, we propose a structured information extraction framework that leverages LLaMA-3.1-70B to extract six semantic dimensions from financial news. Through large-scale experiments on 41,618 news--stock pairs from the FNSPID dataset, we find that (i) FinBERT sentiment features exhibit strong predictive power under nonlinear models (F1\,=\,0.576) but substantially weaker performance under linear models (F1\,=\,0.230), revealing a highly nonlinear sentiment--return relationship; (ii) LLM-extracted structured features, while individually weaker, capture information orthogonal to sentiment, as evidenced by a 53.5\% systematic disagreement rate between the two approaches; and (iii) combining both signal sources yields F1\,=\,0.600, significantly outperforming either alone ($p < 0.0001$), with consistent improvements across all seven event types. Ablation experiments confirm that non-sentiment structural dimensions (event type, impact subject, time horizon, confidence) independently contribute $\Delta\text{F1} = +0.019$ beyond FinBERT alone. Feature importance analysis reveals balanced contributions from all six extracted dimensions (14--21\%), demonstrating that compressing news into a single sentiment score incurs substantial information loss. Our results suggest that the sentiment--semantics decoupling in financial text is systematic and exploitable, opening a new direction for multi-dimensional financial NLP.

\end{abstract}

\section{Introduction}

The relationship between news sentiment and stock returns has been a central topic in both computational finance and natural language processing \cite{tetlock2007giving,loughran2011liability}. The dominant paradigm treats financial text analysis as a sentiment classification problem: given a news article, assign a polarity score (positive, negative, or neutral), and use this score as a feature for downstream prediction tasks. Pre-trained financial language models such as FinBERT \cite{araci2019finbert} have become standard tools in this pipeline, achieving strong performance on sentiment classification benchmarks.

However, this paradigm rests on a reductive assumption: that the market-relevant information in a news article can be adequately summarized by its emotional valence. Consider the headline \textit{``FAA urges airlines to act as wireless carriers plan 5G signal boost.''} FinBERT assigns this a positive sentiment score of 0.94, responding to words like ``boost'' and ``plan.'' Yet the article describes a regulatory warning to airlines---a negative policy event---and the associated stock declined. The surface-level sentiment and the event-level semantics are \emph{decoupled}: the text reads positively, but the event implies negative consequences for the relevant stocks.

This observation motivates our central hypothesis: \textbf{financial news contains multiple information dimensions that are partially orthogonal to surface sentiment, and these dimensions carry independent predictive value for stock price movements.} We identify six such dimensions---sentiment polarity, sentiment intensity, event type, impact subject, time horizon, and extraction confidence---and propose a structured information extraction framework that uses large language models (LLMs) to explicitly decompose financial news along these axes.

Our contributions are threefold:

\begin{enumerate}[leftmargin=2em,itemsep=2pt]
\item \textbf{Sentiment--semantics decoupling hypothesis.} We formalize and quantify the systematic divergence between surface-level sentiment signals and event-level semantic signals in financial news. Across 41,618 samples, FinBERT and LLaMA disagree on sentiment polarity in 53.5\% of cases, with disagreement rates ranging from 39.4\% (merger events) to 67.4\% (uncategorized events), suggesting that the decoupling is not random but structurally related to event complexity.

\item \textbf{Multi-dimensional structured extraction framework.} We design a six-dimensional extraction schema that decomposes financial news into interpretable semantic features using LLaMA-3.1-70B-Instruct. All six dimensions contribute meaningfully to prediction (importance range: 14--21\%), with no single dominant feature, confirming that the framework captures distinct information channels.

\item \textbf{Rigorous complementarity analysis.} Through 1,000-iteration bootstrap experiments with paired statistical tests, we demonstrate that combining FinBERT sentiment with LLM-extracted structured features significantly outperforms either source alone (F1 improvement from 0.576 to 0.600, $p < 0.0001$), with consistent gains across all seven event types (+0.010 to +0.023).
\end{enumerate}

\begin{figure}[t]
\centering
\includegraphics[width=\linewidth]{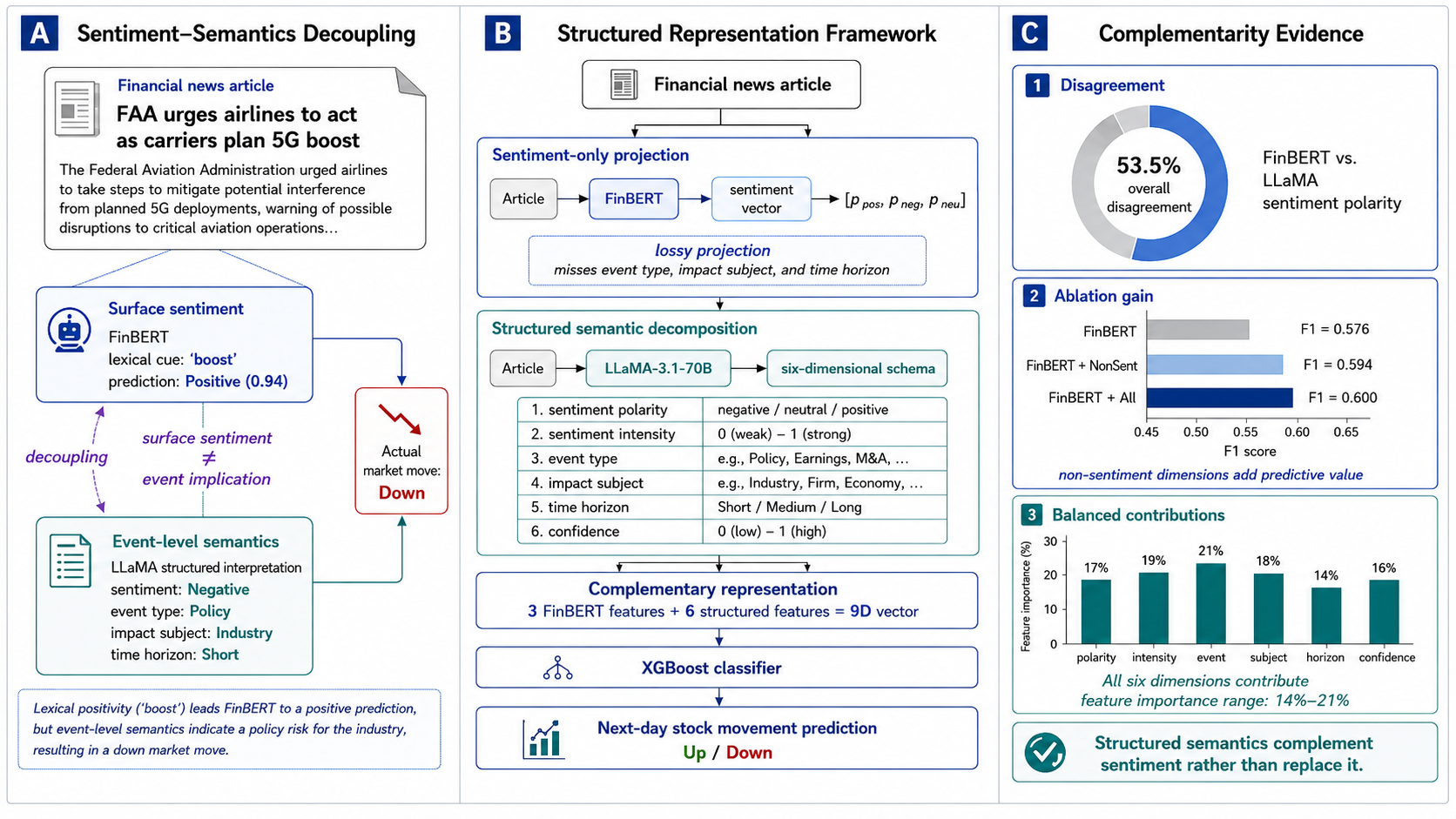}
\caption{
Overview of the proposed framework. 
(A) Sentiment--semantics decoupling: surface lexical sentiment can conflict with event-level market implications. 
(B) Each news article is represented by both FinBERT sentiment probabilities and six LLaMA-extracted semantic dimensions, which are concatenated into a nine-dimensional feature vector for XGBoost-based next-day stock movement prediction. 
(C) Complementarity evidence shows a 53.5\% FinBERT--LLaMA sentiment disagreement rate, improved F1 after adding non-sentiment dimensions, and balanced contributions from all six structured features.
}
\label{fig:framework}
\end{figure}

\section{Related Work}

\paragraph{Financial sentiment analysis.}
The use of textual sentiment for financial prediction has a rich history, beginning with dictionary-based approaches \cite{loughran2011liability,tetlock2007giving} and progressing to machine learning classifiers trained on financial corpora \cite{malo2014good}. The introduction of pre-trained language models marked a significant advance: FinBERT \cite{araci2019finbert}, trained on financial communications, became widely adopted for its strong sentiment classification performance. Huang et al. \cite{huang2023finbert} proposed an alternative FinBERT variant pre-trained on analyst reports. Despite their effectiveness on classification benchmarks, these models produce a single output dimension (sentiment polarity), discarding the rich information structure present in financial text.

\paragraph{LLMs in finance.}
Recent work has explored the application of large language models to financial tasks. Lopez-Lira and Tang \cite{lopezlira2023can} investigated GPT-3.5's ability to predict stock returns from news headlines, finding modest improvements over traditional sentiment approaches. Xie et al. \cite{xie2023wallstreetneophyte} benchmarked ChatGPT on financial sentiment classification. BloombergGPT \cite{wu2023bloomberggpt} demonstrated the value of domain-specific pre-training. However, most studies use LLMs as improved sentiment classifiers rather than as structured information extractors, leaving the multi-dimensional information content of financial text unexplored.

\paragraph{News-based stock prediction.}
Predicting stock movements from news has been approached through diverse methodologies, including event-driven models \cite{ding2015deep}, attention-based architectures \cite{hu2018listening}, and graph neural networks for modeling inter-stock relationships \cite{feng2019temporal}. Jiang and Zeng \cite{xie2023wallstreetneophyte} achieved accuracy in the 51--56\% range on next-day prediction tasks, consistent with the efficient market hypothesis's prediction that news-based signals should be weak but nonzero. Our work does not aim to push absolute prediction accuracy beyond this established range; rather, we seek to understand \emph{which information dimensions in financial news carry predictive signal}, a question that accuracy alone cannot answer.

\paragraph{Structured information extraction.}
Information extraction from text has a long history in NLP \cite{jurafsky2023speech}, but its application to financial prediction remains limited. Jacobs and Hoste \cite{jacobs2020extracting} extracted economic events but did not analyze dimensional contributions. Recent work on tool-augmented LLMs \cite{schick2024toolformer} suggests that structured extraction can complement end-to-end approaches. Our framework differs in that it explicitly decomposes financial text into \emph{multiple interpretable dimensions} and analyzes the predictive contribution of each.

\section{Methodology}

\subsection{Problem Formulation}

Let $\mathcal{D} = \{(a_i, s_i, t_i, y_i)\}_{i=1}^{N}$ denote a dataset of financial news articles, where $a_i$ is the article text, $s_i$ is the associated stock symbol, $t_i$ is the publication date, and $y_i \in \{0, 1\}$ is the binary price movement label (1 if the closing price on the next trading day exceeds the closing price on day $t_i$, 0 otherwise). Our goal is to learn a mapping $f: \mathcal{X} \rightarrow \{0, 1\}$ from extracted features $\mathbf{x}_i \in \mathcal{X}$ to price movement labels.

We consider three feature representations: (i) sentiment features $\mathbf{x}_i^{\text{sent}} \in \mathbb{R}^3$ from FinBERT, (ii) structured features $\mathbf{x}_i^{\text{struct}} \in \mathbb{R}^6$ from LLM extraction, and (iii) the concatenation $\mathbf{x}_i^{\text{all}} = [\mathbf{x}_i^{\text{sent}}; \mathbf{x}_i^{\text{struct}}] \in \mathbb{R}^9$.

The standard FinBERT pipeline defines a feature extraction function $\phi^{\text{sent}}: \mathcal{A} \rightarrow \Delta^2$, mapping each article to a point on the 2-simplex. This projection is inherently lossy: all market-relevant information in $a_i$ is compressed onto a two-dimensional manifold. Our central hypothesis is that there exist information dimensions orthogonal to $\mathbf{x}_i^{\text{sent}}$ that carry independent predictive value for $y_i$. Formally, we hypothesize:
\begin{equation}
I\!\left(\mathbf{x}_i^{\text{struct}};\, y_i \mid \mathbf{x}_i^{\text{sent}}\right) > 0
\label{eq:hypothesis}
\end{equation}
where $I(\cdot\,;\,\cdot \mid \cdot)$ denotes conditional mutual information. If structured features improve prediction beyond sentiment alone, the hypothesis holds and the two pipelines are complementary.

\subsection{Dataset}

We use the FNSPID dataset \cite{dong2024fnspid}, a large-scale collection of 15.7 million time-aligned financial news articles for S\&P 500 companies spanning 1999--2023. We focus on the NASDAQ news subset and apply the following preprocessing pipeline:

\begin{enumerate}[leftmargin=2em,itemsep=2pt]
\item \textbf{Quality filtering.} Remove articles with invalid stock symbols or text length below 200 characters, yielding approximately 2.4 million valid records.
\item \textbf{Temporal and coverage filtering.} Restrict to January 2019--December 2023, avoiding pre-crisis structural breaks while covering multiple market regimes. Select the 100 most frequently covered stocks to ensure sufficient per-ticker sample density.
\item \textbf{Balanced sampling.} Sample up to 500 articles per stock uniformly across the date range, preventing high-coverage tickers from dominating the corpus. This yields 50,000 articles across 100 stocks.
\item \textbf{Label construction.} For each article published on date $t_i$ for stock $s_i$:
\begin{equation}
y_i = \mathbf{1}\!\left[\text{close}_{s_i,\, t_i^+} > \text{close}_{s_i,\, t_i}\right]
\end{equation}
where $t_i^+$ is the next trading day. Close-to-close returns are used to avoid confounding intraday price impact. After price alignment: $N = 41{,}618$, positive rate 48.7\%.
\end{enumerate}

\subsection{FinBERT Sentiment Features}

For each article $a_i$, we extract sentiment features using ProsusAI/FinBERT \cite{araci2019finbert}, a BERT model fine-tuned on the Financial PhraseBank corpus. The model outputs a probability distribution over three classes:
\begin{equation}
\mathbf{x}_i^{\text{sent}} = \phi^{\text{sent}}(a_i) = \left(p_i^{\text{pos}},\, p_i^{\text{neg}},\, p_i^{\text{neu}}\right) \in \Delta^2
\end{equation}
where $\Delta^2 = \{(p,q,r) : p+q+r=1,\; p,q,r \geq 0\}$ denotes the 2-simplex. For articles exceeding the 512-token limit, we truncate to the first 512 tokens; this truncation is a source of information asymmetry relative to LLaMA, which we discuss in §\ref{sec:discussion}.

\subsection{Structured Information Extraction Framework}

\paragraph{Dimension design.}
The six dimensions decompose financial news along three axes: \emph{sentiment} ($z^{(1)}, z^{(2)}$: direction and intensity), \emph{event} ($z^{(3)}, z^{(4)}$: type and impact scope), and \emph{temporal} ($z^{(5)}, z^{(6)}$: horizon and confidence)---corresponding to the analyst questions \emph{what reaction?}, \emph{what happened and to whom?}, and \emph{when and how clearly?}

We define $\phi^{\text{struct}}: \mathcal{A} \rightarrow \mathcal{Z}$ with output space $\mathcal{Z} = \mathcal{S} \times [-1,1] \times \mathcal{E} \times \mathcal{C} \times \mathcal{H} \times [0,1]$, where $\mathcal{S} = \{\texttt{pos}, \texttt{neg}, \texttt{neu}\}$, $\mathcal{E} = \{\texttt{earnings}, \texttt{merger}, \texttt{policy}, \texttt{product}, \texttt{mgmt}, \texttt{macro}, \texttt{other}\}$, $\mathcal{C} = \{\texttt{company}, \texttt{industry}, \texttt{macro}\}$, $\mathcal{H} = \{\texttt{short}, \texttt{long}\}$. Dimensions $z^{(3)}$--$z^{(5)}$ are entirely absent from $\phi^{\text{sent}}$'s output space, making them the primary source of complementary information.

\paragraph{Feature encoding.}
Raw output $\mathbf{z}_i \in \mathcal{Z}$ is mapped to $\mathbb{R}^6$ (XGBoost, integer encoding $\psi^{\text{int}}$) or $\mathbb{R}^{d}$ (LR, one-hot encoding $\psi^{\text{oh}}$). One-hot encoding prevents LR from treating nominal categories as ordinal; XGBoost is less sensitive to this. Both encodings yield equivalent predictive performance ($\Delta\text{F1} = 0.0002$, Appendix~\ref{app:encoding}).

\paragraph{Zero-shot extraction.}
We use LLaMA-3.1-70B-Instruct \cite{dubey2024llama3} as a zero-shot extractor. Supervised alternatives are impractical given the absence of annotated training data for these dimensions, and LLMs handle context-dependent judgments beyond the reach of lexical classifiers. Articles are truncated to 2,000 characters (Appendix~\ref{app:prompt}); greedy decoding is used. Parse success rate: $|\mathcal{V}| = 41{,}044 / 41{,}618$ (98.6\%); failures excluded.

\subsection{Prediction Models}

LR and XGBoost serve distinct analytical functions: LR tests whether features carry \emph{linearly separable} predictive signal, while XGBoost quantifies the \emph{maximum extractable} predictive value and provides split-based feature importance for measuring dimensional contributions. We deliberately avoid deep learning models, whose goal is maximizing absolute performance rather than isolating feature contributions.

\paragraph{Logistic regression (LR).} $L_2$-regularized linear classifier with default hyperparameters. Categorical features encoded via $\psi^{\text{oh}}$.

\paragraph{XGBoost.} Gradient-boosted tree ensemble \cite{chen2016xgboost} with 100 estimators and log-loss objective. Categorical features encoded via $\psi^{\text{int}}$.

\subsection{Evaluation Protocol}

Single train--test splits in financial prediction yield highly variable results, and comparing models evaluated on different splits conflates data-partition effects with model differences. Our bootstrap protocol addresses both problems simultaneously.

In each of $B = 1{,}000$ iterations, we draw a bootstrap resample of size $N$ with replacement, generate a shared random permutation $\pi_b$, and split into $80\%$ training and $20\%$ test sets. All configurations are evaluated on the \emph{identical} partition within each iteration, enabling valid paired comparisons via:
\begin{equation}
d_b^{(A,B)} = \text{F1}_b^{(A)} - \text{F1}_b^{(B)}, \quad T^{(A,B)} = \frac{\bar{d}^{(A,B)}}{s_{d^{(A,B)}} / \sqrt{B}}
\end{equation}
For $K$ pairwise comparisons we apply Bonferroni correction with $\alpha' = \alpha/K$, $\alpha = 0.05$.

\section{Experiments}

\subsection{Experimental Setup}

All experiments use 1,000-iteration bootstrap evaluation. In each iteration, we (i) draw a bootstrap sample of size $N$ with replacement from the full dataset, (ii) generate a random permutation of indices and split into 80\% training / 20\% test sets, and (iii) train and evaluate each model on the \emph{identical} train--test partition to ensure fair comparison. This design addresses two common pitfalls: single-split instability and inconsistent partitions across compared models.

We report mean $\pm$ standard deviation across bootstrap iterations for accuracy, F1 score, and AUROC. Statistical significance is assessed via paired $t$-tests on per-iteration F1 scores, with the Bonferroni correction applied for multiple comparisons.

All experiments were conducted on NVIDIA A100 80GB GPUs using XGBoost's CUDA-accelerated histogram method.

\subsection{Main Results}

Table~\ref{tab:main} presents the main experimental results. Several findings emerge.

\begin{table}[t]
\centering
\caption{Main experimental results (bootstrap, $B = 1{,}000$ iterations). $^\dagger$\,indicates $p < 0.0001$ vs.\ \textsc{All+XGB} (paired $t$-test on F1). LR configurations use one-hot encoding for categorical features.}
\label{tab:main}
\small
\begin{tabular}{@{}llccc@{}}
\toprule
\textbf{Features} & \textbf{Model} & \textbf{Accuracy} & \textbf{F1} & \textbf{AUROC} \\
\midrule
---                & Random & $0.500 \pm 0.006^\dagger$ & $0.488 \pm 0.007^\dagger$ & $0.500 \pm 0.000^\dagger$ \\
\midrule
\multirow{2}{*}{\fb~sentiment}
& LR      & $0.512 \pm 0.006^\dagger$ & $0.230 \pm 0.079^\dagger$ & $0.502 \pm 0.006^\dagger$ \\
& XGBoost & $0.602 \pm 0.006^\dagger$ & $0.576 \pm 0.009^\dagger$ & $0.526 \pm 0.006^\dagger$ \\
\midrule
Sentiment diff ($p^+\!-\!p^-$)
& LR      & --- & $0.000 \pm 0.000^\dagger$ & --- \\
& XGBoost & --- & $0.435 \pm 0.014^\dagger$ & --- \\
\midrule
\multirow{2}{*}{LLM structured}
& LR      & $0.516 \pm 0.006^\dagger$ & $0.330 \pm 0.032^\dagger$ & $0.515 \pm 0.006^\dagger$ \\
& XGBoost & $0.527 \pm 0.006^\dagger$ & $0.450 \pm 0.027^\dagger$ & $0.507 \pm 0.005^\dagger$ \\
\midrule
Combined (\fb~+ LLM)
& LR      & --- & $0.346 \pm 0.033^\dagger$ & $0.515 \pm 0.006^\dagger$ \\
& XGBoost & $\mathbf{0.623 \pm 0.006}$ & $\mathbf{0.600 \pm 0.009}$ & $\mathbf{0.528 \pm 0.006}$ \\

\bottomrule
\end{tabular}
\end{table}

\paragraph{Finding 1: The sentiment--return relationship is highly nonlinear.} FinBERT features under logistic regression yield F1\,=\,0.230, substantially below the XGBoost counterpart (0.576) and well below the random baseline (0.488). The same features under XGBoost achieve a $2.5\times$ improvement. To confirm this reflects genuine nonlinearity rather than an implementation artifact, we test a single-feature linear baseline: the sentiment difference $p^+ - p^-$. Under logistic regression this yields F1\,=\,0.000; under XGBoost it yields F1\,=\,0.435 ($p < 0.0001$). The identical feature produces near-zero performance under a linear model and meaningful performance under a nonlinear one, directly establishing that the sentiment--return mapping is fundamentally nonlinear.

\paragraph{Finding 2: LLM features are individually weaker but complementary.} LLM-extracted features under XGBoost achieve F1\,=\,0.450, substantially below FinBERT+XGB (0.576). However, combining both feature sets yields F1\,=\,0.600, significantly higher than either source alone ($p < 0.0001$ for both comparisons). This pattern---weak individually, strong in combination---is the hallmark of complementary information sources.

\paragraph{Finding 2a: Non-sentiment dimensions drive the complementarity gain.} To isolate whether the improvement stems from the LLM's sentiment judgment or from the structural dimensions (event type, impact subject, time horizon, confidence), we ablate the two sentiment dimensions from the LLM feature set. \textsc{FinBERT+NonSent} achieves F1\,=\,0.594, significantly outperforming FinBERT alone ($\Delta = +0.019$, $p < 0.0001$). The LLM sentiment dimensions contribute an additional $\Delta = +0.006$ on top of this ($p < 0.0001$). Both layers of complementarity are statistically significant, confirming that structural features carry predictive information independent of any sentiment signal.

\paragraph{Finding 3: LLM features are more linearly accessible.} Under logistic regression with one-hot encoding, LLM features (F1\,=\,0.330) outperform FinBERT features (F1\,=\,0.230), despite being weaker under XGBoost. This advantage is robust to encoding choice: ordinal encoding yields F1\,=\,0.305, and one-hot encoding raises this to 0.330 ($\Delta = +0.025$, $p < 0.0001$), confirming it is not an artifact of categorical encoding. The result suggests that the six-dimensional structured representation encodes predictive information in a more linearly separable form than the three-dimensional sentiment probability simplex.

\paragraph{Contextualizing absolute performance.} The accuracy range of 50--62\% is consistent with prior work on news-based next-day prediction. Lopez-Lira and Tang \cite{lopezlira2023can} report accuracy in the 51--56\% range. Our contribution is not to push absolute performance, but to decompose the information structure of financial news and demonstrate that sentiment is insufficient.

\subsection{Feature Importance Analysis}

Table~\ref{tab:importance} reports the bootstrap-averaged feature importance scores for the six LLM-extracted dimensions in the LLaMA+XGB configuration.

\begin{table}[t]
\centering
\caption{Feature importance of LLM-extracted dimensions in \allxgb~(bootstrap mean $\pm$ std, $B = 1{,}000$). Importance scores are XGBoost split-based gain under integer encoding, reflecting each feature's utility in tree partitioning. All six dimensions contribute meaningfully, with no single dominant feature (range: 9.8--21.3\%).}
\label{tab:importance}
\small
\begin{tabular}{@{}lc@{}}
\toprule
\textbf{Feature} & \textbf{Importance} \\
\midrule
Sentiment score    & $0.213 \pm 0.020$ \\
Sentiment (encoded)  & $0.174 \pm 0.034$ \\
Event type (encoded)   & $0.165 \pm 0.015$ \\
Confidence           & $0.153 \pm 0.019$ \\
Time horizon (encoded) & $0.151 \pm 0.017$ \\
Impact subject (encoded) & $0.144 \pm 0.016$ \\
\bottomrule
\end{tabular}
\end{table}

The importance distribution is notably balanced (range: 14.4--21.3\%), confirming that no single dimension dominates and that the framework captures six distinct information channels. Two observations are particularly noteworthy:

\textit{Event type matters.} The event type dimension (16.5\%) contributes comparably to the sentiment score (21.3\%), despite being entirely absent from FinBERT's output space. This directly supports our hypothesis that sentiment-only approaches incur information loss by discarding event semantics.

\textit{Confidence is informative.} The model's self-assessed extraction confidence (15.3\%) carries substantial predictive signal. We interpret this as reflecting article clarity: articles with ambiguous implications (low confidence) may correspond to uncertain market reactions, making the confidence score an implicit proxy for event interpretability.

\subsection{Event-Type Conditioned Analysis}

Table~\ref{tab:event} presents F1 scores stratified by event type, comparing FinBERT-only, LLaMA-only, and combined features, all using XGBoost.

\begin{table}[t]
\centering
\caption{F1 scores by event type (bootstrap, $B = 1{,}000$). Combined features (\textsc{All}) consistently outperform \fb~alone across all event types. $\Delta$ denotes the \textsc{All} vs.\ \fb~improvement.}
\label{tab:event}
\small
\begin{tabular}{@{}lrcccr@{}}
\toprule
\textbf{Event} & $n$ & \textbf{\fb} & \textbf{\llm} & \textbf{All} & $\Delta$ \\
\midrule
Merger     & 1{,}207  & $.728 \pm .035$ & $.455 \pm .085$ & $.738 \pm .034$ & $+.010$ \\
Policy     & 1{,}601  & $.720 \pm .031$ & $.494 \pm .056$ & $.739 \pm .030$ & $+.019$ \\
Management & 1{,}792  & $.706 \pm .030$ & $.404 \pm .080$ & $.725 \pm .030$ & $+.019$ \\
Macro      & 9{,}630  & $.673 \pm .014$ & $.521 \pm .035$ & $.696 \pm .013$ & $+.023$ \\
Product    & 6{,}752  & $.661 \pm .017$ & $.450 \pm .086$ & $.678 \pm .016$ & $+.017$ \\
Other      & 12{,}788 & $.642 \pm .013$ & $.450 \pm .056$ & $.658 \pm .013$ & $+.016$ \\
Earnings   & 7{,}050  & $.636 \pm .017$ & $.304 \pm .070$ & $.656 \pm .017$ & $+.020$ \\
\bottomrule
\end{tabular}
\end{table}

\paragraph{Consistent complementarity.} The combined model outperforms FinBERT alone on all seven event types, with improvements ranging from +0.010 (merger) to +0.023 (macro). This consistency is important: it demonstrates that LLM-extracted features provide genuine complementary information rather than benefiting from a specific event subtype.

\paragraph{FinBERT dominates individual performance.} LLaMA features alone underperform FinBERT on every event type, with gaps ranging from 0.15 (macro) to 0.33 (earnings). The earnings category shows the largest deficit, likely because earnings reports use standardized financial language where FinBERT's domain-specific training provides a strong advantage. Conversely, macro events---which often involve complex policy implications---show the smallest gap, suggesting that LLM-extracted semantic features are relatively more valuable for events requiring deeper contextual understanding.

\paragraph{Largest complementary gains on macro events.} The +0.023 improvement on macro events (the largest category at $n = 9{,}630$) is the most impactful result, as it affects the most predictions and corresponds to the event type where sentiment and semantics are most likely to diverge (e.g., policy announcements with positive framing but negative market implications).

\section{Analysis}

\subsection{Quantifying Sentiment--Semantics Decoupling}

To directly test our central hypothesis, we measure the disagreement rate between FinBERT's sentiment classification and LLaMA's sentiment assessment. For each sample, we binarize both predictions (positive vs.\ non-positive) and compute the fraction of samples where the two models disagree.

\begin{figure}[t]
\centering
\includegraphics[width=\linewidth]{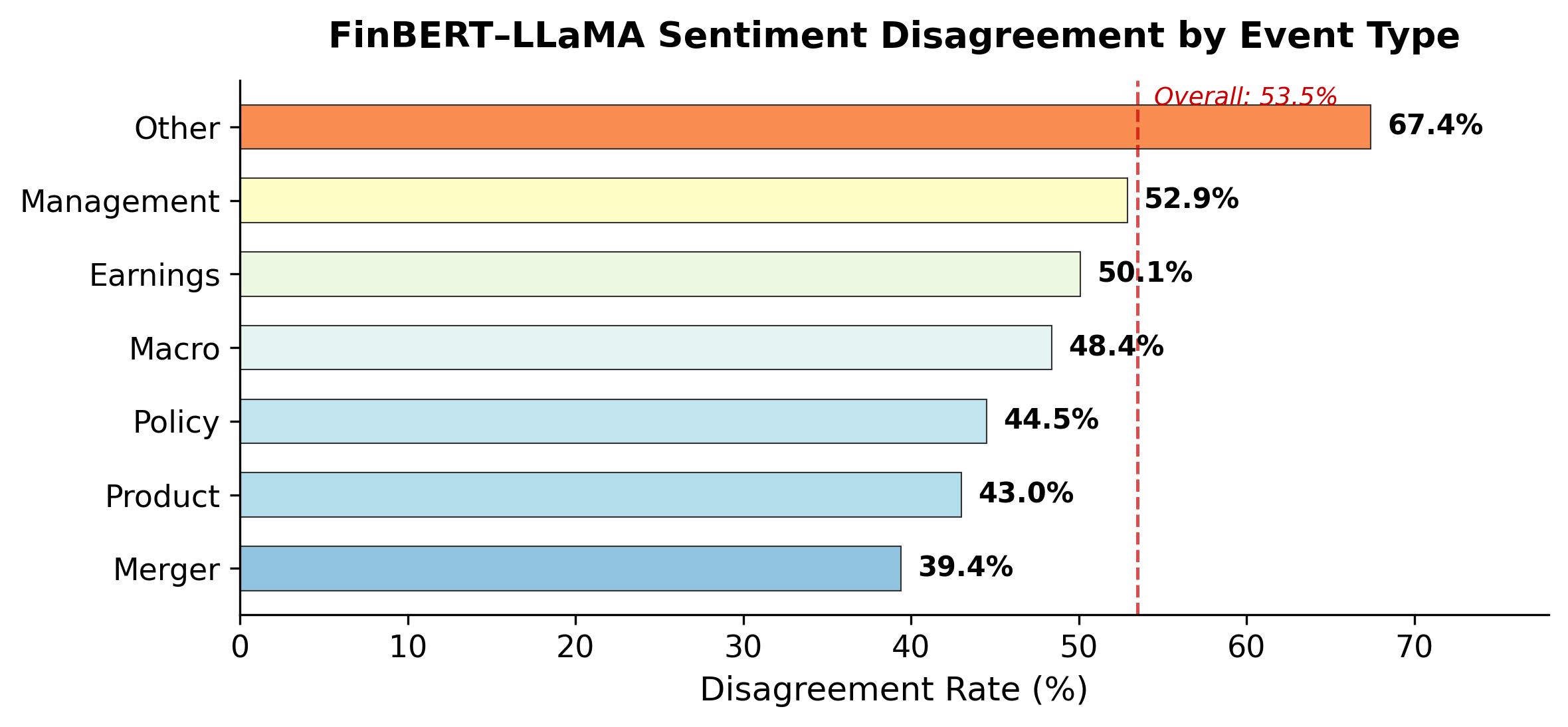}
\caption{FinBERT--LLaMA sentiment disagreement rate by event type. Higher disagreement indicates greater decoupling between surface-level lexical sentiment and event-level semantics. Merger events, with their unambiguous transactional language, show the lowest disagreement; uncategorized events show the highest.}
\label{fig:disagreement}
\end{figure}

\begin{table}[t]
\centering
\caption{Sentiment disagreement rate between \fb~and \llm, stratified by event type (see also Figure~\ref{fig:disagreement}). Higher disagreement indicates greater decoupling between surface sentiment and event semantics.}
\label{tab:disagree}
\small
\begin{tabular}{@{}lcc@{}}
\toprule
\textbf{Event Type} & \textbf{Disagreement Rate} & \textbf{Interpretation} \\
\midrule
Other      & 67.4\% & Highest: semantically ambiguous events \\
Management & 52.9\% & \\
Earnings   & 50.1\% & \\
Macro      & 48.4\% & \\
Policy     & 44.5\% & \\
Product    & 43.0\% & \\
Merger     & 39.4\% & Lowest: semantically unambiguous events \\
\midrule
\textbf{Overall} & \textbf{53.5\%} & \\
\bottomrule
\end{tabular}
\end{table}

The overall disagreement rate of 53.5\% (Table~\ref{tab:disagree}) is striking: in more than half of all articles, FinBERT and LLaMA reach opposite sentiment conclusions. Moreover, the disagreement pattern is structurally meaningful:

\textit{Semantically complex events show higher decoupling.} The ``other'' category (catch-all for events that resist clean categorization) has the highest disagreement rate at 67.4\%, while merger events---which typically involve clear transactional language---have the lowest at 39.4\%. This gradient suggests that the decoupling is not random noise but reflects genuine differences in how surface-level lexical cues and event-level semantics align across event types.

\textit{Decoupling does not predict market direction.} Among samples where the two models disagree, 48.9\% are positive (up) labels; among agreeing samples, 48.5\% are positive. The near-identical label distributions rule out a simple interpretation where one model is ``right'' and the other ``wrong.'' Instead, both capture \emph{different aspects} of the text's informativeness, consistent with our complementarity finding.

\subsection{Why FinBERT Fails Under Linear Models}

The substantially reduced performance of FinBERT+LR (F1\,=\,0.230, std\,=\,0.079) compared to FinBERT+XGB (F1\,=\,0.576) deserves careful analysis, as it reveals a fundamental property of sentiment signals in financial prediction.

To rule out implementation artifacts, we test the simplest possible sentiment signal: the scalar difference $p^+ - p^-$, which collapses FinBERT's three-dimensional output to a single signed value. Under logistic regression, this yields F1\,=\,0.000 with zero variance---the model predicts a single class in every bootstrap iteration. Under XGBoost, the same feature yields F1\,=\,0.435 ($p < 0.0001$). This controlled comparison establishes that the failure is not a threshold artifact or preprocessing issue, but a genuine property of the sentiment--return relationship: the same information that XGBoost can exploit is largely inaccessible to linear models.

We hypothesize two contributing factors. First, the relationship is \emph{context-dependent}: positive sentiment in an earnings report predicts different outcomes than positive sentiment in a policy announcement. This interaction effect cannot be captured by a linear function of sentiment features alone, but can be captured by tree-based models that partition the feature space. Second, the simplex constraint reduces FinBERT's effective dimensionality to a 2D manifold, limiting linear separability.

The high variance of FinBERT+LR (std\,=\,0.079) further indicates instability: in some iterations the linear model collapses to single-class prediction, while in others it achieves modest F1. This contrasts sharply with XGBoost's low variance (std\,=\,0.009).

\subsection{Case Studies}

To provide intuition for the sentiment--semantics decoupling, we present representative examples where FinBERT and LLaMA disagree and the event-level interpretation proves more aligned with market outcomes.

\begin{table}[t]
\centering
\caption{Representative cases of sentiment--semantics decoupling. FinBERT's lexical sentiment conflicts with LLaMA's event-level assessment, and the event-level interpretation aligns with the actual market outcome.}
\label{tab:cases}
\small
\begin{tabular}{@{}p{4.2cm}ccc@{}}
\toprule
\textbf{Headline} & \textbf{\fb} & \textbf{\llm} & \textbf{Actual} \\
\midrule
\textit{FAA urges airlines to act as carriers plan 5G boost} & pos (0.94) & neg / policy & Down \\[3pt]
\textit{Australian coal royalty hike could nudge others} & pos (0.94) & neg / policy & Down \\[3pt]
\textit{Anglo American plans \$1.8B capex cuts by 2026} & pos (0.87) & neg / mgmt & Down \\
\midrule
\textit{US ethanol expands to lower-carbon aviation} & neg (0.96) & pos / policy & Up \\[3pt]
\textit{Anglo American's Los Bronces gets env.\ permit} & neg (0.93) & pos / policy & Up \\[3pt]
\textit{5 Hot Airline Stocks Ready for Takeoff} & neg (0.92) & pos / earnings & Up \\
\bottomrule
\end{tabular}
\end{table}

These cases (Table~\ref{tab:cases}) illustrate a recurring pattern: FinBERT responds to \emph{lexical sentiment cues} (``boost,'' ``cuts,'' ``lower-carbon'') while LLaMA interprets the \emph{event implications} (regulatory warning, cost reduction pressure, industry expansion). The word ``cuts'' triggers positive sentiment in FinBERT (perhaps associated with cost-cutting efficiency), but LLaMA recognizes \$1.8B capital expenditure reduction as a negative management signal. Similarly, ``lower-carbon'' triggers negative sentiment (perhaps associated with restrictions), but LLaMA identifies a sector expansion opportunity.

While these cases are selected to illustrate the phenomenon rather than to provide statistical evidence, they are consistent with the quantitative findings: the two models extract \emph{qualitatively different} information from the same text, and the combination of both provides a richer representation than either alone.

\section{Discussion}
\label{sec:discussion}

\paragraph{Implications for financial NLP.}
Our results challenge the prevailing practice of using sentiment as the sole NLP signal for financial prediction. The 53.5\% disagreement rate demonstrates that surface sentiment and event semantics are \emph{partially orthogonal} information dimensions. This suggests that the field should move toward multi-dimensional text representations that explicitly model event type, impact scope, and temporal dynamics, rather than treating all financial text analysis as a sentiment classification problem.

\paragraph{Why not use the LLM directly?}
A natural question is why we extract structured features rather than fine-tuning the LLM end-to-end for stock prediction. We offer two reasons. First, interpretability: our framework reveals \emph{which} dimensions carry signal (all six, balanced at 14--21\%), a finding that would be opaque in an end-to-end model. Second, practical cost: structured extraction requires a single inference pass per article, after which the features can be reused across arbitrary downstream models and time horizons. End-to-end fine-tuning would require retraining for each prediction task specification.

\paragraph{Limitations.}
Several limitations warrant acknowledgment. (i) \textit{Pretraining data overlap}: LLaMA-3.1-70B was trained on data with a cutoff that overlaps our 2019--2023 evaluation period, raising the possibility that the model has implicit knowledge of post-publication outcomes. However, our framework extracts structural semantic dimensions (event type, impact scope, time horizon) rather than factual predictions, and the same concern applies equally to any modern LLM used in financial NLP research. We treat this as a shared limitation of the field rather than a confound specific to our approach. (ii) \textit{Bootstrap overlap}: bootstrap sampling with replacement may place duplicate instances of the same original sample in both the training and test sets within a single iteration, which can optimistically bias absolute performance estimates. Our bootstrap protocol is primarily used for variance estimation and paired comparisons between feature sets evaluated under identical conditions; relative conclusions remain valid. (iii) \textit{Temporal validity}: our bootstrap protocol uses random rather than temporally-ordered splits, which does not enforce causal ordering. A walk-forward evaluation would provide more conservative absolute performance estimates for deployment scenarios, though the relative comparisons between feature sets remain internally valid under identical splits. (iii) \textit{Extraction quality}: we have no ground truth for the correctness of LLM extractions---the model may systematically misclassify certain event types. The 98.6\% parse success rate is an indirect quality indicator. (v) \textit{Confidence signal}: the LLM's self-assessed confidence is uncalibrated and may partially proxy for article length or lexical complexity rather than genuine semantic clarity. (vi) \textit{Categorical encoding}: integer encoding of categorical features imposes arbitrary ordinal structure on nominal dimensions such as event type; we report one-hot XGBoost results in the Appendix where prediction performance is equivalent (F1: $0.487 \pm 0.007$ vs $0.487 \pm 0.007$, $\Delta = 0.0002$), confirming predictive conclusions are robust to encoding choice, though feature importance scores differ. (vii) \textit{Context length asymmetry}: FinBERT truncates input to 512 tokens while LLaMA processes up to 2,000 characters, introducing an asymmetry in available context that may partially confound comparisons of what each approach captures. (viii) \textit{Sample scope}: our analysis covers 100 NASDAQ-listed stocks from 2019--2023. Generalization to other markets, asset classes, or time periods remains untested. (ix) \textit{Single LLM}: we use only LLaMA-3.1-70B-Instruct; cross-LLM robustness is not evaluated.

\paragraph{Broader implications.}
The sentiment--semantics decoupling we observe is likely not unique to financial text. Medical news (``breakthrough treatment shows severe side effects''), political reporting (``controversial bill gains bipartisan support''), and legal documents all involve texts where surface sentiment diverges from domain-specific implications. Our structured extraction framework could be adapted to these domains, potentially revealing similar complementarity patterns.

\section{Conclusion}

We proposed and validated the hypothesis that financial news contains multi-dimensional information that surface-level sentiment analysis systematically fails to capture. Through a structured extraction framework that decomposes financial text into six semantic dimensions using LLaMA-3.1-70B, we demonstrated three key findings: (1) the sentiment--return relationship is highly nonlinear, rendering sentiment features useless under linear models; (2) FinBERT and LLaMA exhibit a 53.5\% systematic disagreement rate, confirming that surface sentiment and event semantics are partially decoupled; and (3) combining both signal sources yields consistent improvements across all event types, with the combined model significantly outperforming either source alone.

Our work opens several directions for future research. The balanced importance of all six extracted dimensions (14--21\%) suggests that even richer extraction schemas---incorporating, for example, named entities, causal relationships, or market expectations---could yield further improvements. The event-type conditioning analysis suggests that adaptive models, which weight different signal sources based on event characteristics, may capture the complementarity more effectively than simple feature concatenation. Finally, extending this framework to multi-day prediction horizons, cross-market settings, and other domains with sentiment--semantics decoupling would test the generality of our findings.

\bibliographystyle{splncs04}

\newpage
\appendix
\section{Appendix}

\subsection{LLM Extraction Distribution}

Table~\ref{tab:dist} provides the full distribution of LLM-extracted features.

\begin{table}[h]
\centering
\caption{Distribution of LLM-extracted features across 41,618 samples.}
\label{tab:dist}
\small
\begin{tabular}{@{}llrr@{}}
\toprule
\textbf{Dimension} & \textbf{Value} & \textbf{Count} & \textbf{\%} \\
\midrule
\multirow{3}{*}{Sentiment}
& Positive & 23{,}863 & 57.3 \\
& Neutral  & 12{,}257 & 29.4 \\
& Negative &  4{,}924 & 11.8 \\
\midrule
\multirow{7}{*}{Event type}
& Other      & 12{,}788 & 30.7 \\
& Macro      &  9{,}630 & 23.1 \\
& Earnings   &  7{,}050 & 16.9 \\
& Product    &  6{,}752 & 16.2 \\
& Management &  1{,}792 &  4.3 \\
& Policy     &  1{,}601 &  3.8 \\
& Merger     &  1{,}207 &  2.9 \\
\bottomrule
\end{tabular}
\end{table}

\subsection{Product vs.\ Earnings Signal Characteristics}

Table~\ref{tab:prodvsear} compares signal properties between product and earnings events, the two largest non-macro categories.

\begin{table}[h]
\centering
\caption{Comparison of signal characteristics between product and earnings events.}
\label{tab:prodvsear}
\small
\begin{tabular}{@{}lcc@{}}
\toprule
\textbf{Metric} & \textbf{Product} & \textbf{Earnings} \\
\midrule
Sample size          & 6{,}752 & 7{,}050 \\
\fb~confidence (mean)  & 0.820   & 0.849 \\
Disagreement rate    & 43.0\%  & 50.1\% \\
Avg.\ article length & 4{,}832 chars & 5{,}338 chars \\
\fb~positive (mean)    & 0.328   & 0.433 \\
LLM sentiment score (mean) & 0.445 & 0.338 \\
All vs.\ \fb~$\Delta$F1 & +0.017  & +0.020 \\
\bottomrule
\end{tabular}
\end{table}

The earnings category exhibits higher FinBERT confidence (0.849 vs.\ 0.820) and higher disagreement (50.1\% vs.\ 43.0\%) simultaneously. This apparent paradox resolves when we recognize that FinBERT is \emph{more confident but more often wrong} on earnings articles: the standardized financial language in earnings reports triggers strong but potentially misleading sentiment signals.

\subsection{Bootstrap Methodology Details}

Our bootstrap procedure addresses the well-documented instability of single train--test splits in financial prediction tasks. The critical design choice is the use of a shared permutation-based split within each bootstrap iteration:

\begin{algorithmic}[1]
\FOR{$b = 1$ to $B = 1{,}000$}
  \STATE $\mathcal{I}_b \leftarrow$ sample $N$ indices with replacement from $\{1, \ldots, N\}$
  \STATE $\pi_b \leftarrow$ random permutation of $\{1, \ldots, N\}$
  \STATE $\mathcal{T}_b^{\text{train}} \leftarrow \{\pi_b[1], \ldots, \pi_b[\lfloor 0.8N \rfloor]\}$
  \STATE $\mathcal{T}_b^{\text{test}} \leftarrow \{\pi_b[\lfloor 0.8N \rfloor + 1], \ldots, \pi_b[N]\}$
  \FOR{each model configuration $m$}
    \STATE Train $m$ on $\{(\mathbf{x}_{i}^{(m)}, y_i) : i \in \mathcal{I}_b[\mathcal{T}_b^{\text{train}}]\}$
    \STATE Evaluate on $\{(\mathbf{x}_{i}^{(m)}, y_i) : i \in \mathcal{I}_b[\mathcal{T}_b^{\text{test}}]\}$
  \ENDFOR
\ENDFOR
\end{algorithmic}

This ensures that all model comparisons within each iteration use identical training and test samples, enabling valid paired statistical tests. An earlier version of our code used \texttt{train\_test\_split} with \texttt{random\_state=None}, which generated different splits for different models within the same iteration, leading to inflated or deflated comparison statistics. All results reported in this paper use the corrected permutation-based procedure.

\subsection{Encoding Robustness: One-Hot vs.\ Integer Encoding for XGBoost}
\label{app:encoding}

To verify that the main results are not artifacts of integer encoding, we replicate the \allxgb~configuration using one-hot encoding for all categorical features. Table~\ref{tab:encoding} reports the comparison.

\begin{table}[h]
\centering
\caption{Prediction performance under integer vs.\ one-hot encoding for XGBoost ($B=1{,}000$).}
\label{tab:encoding}
\small
\begin{tabular}{@{}lcc@{}}
\toprule
\textbf{Encoding} & \textbf{F1} & \textbf{AUROC} \\
\midrule
Integer (main paper) & $0.487 \pm 0.007$ & $0.528 \pm 0.006$ \\
One-hot              & $0.487 \pm 0.007$ & $0.528 \pm 0.006$ \\
\midrule
$\Delta$ & $+0.0002$ & $+0.0001$ \\
\bottomrule
\end{tabular}
\end{table}

Prediction performance is essentially identical across encoding schemes ($\Delta\text{F1} = 0.0002$), confirming that the complementarity conclusions are not encoding artifacts.

\subsection{LLM Extraction Prompt}
\label{app:prompt}

The following prompt template was used for all LLaMA-3.1-70B-Instruct extractions. Articles were truncated to 2,000 characters prior to insertion. Inference used greedy decoding (temperature\,=\,0, \texttt{do\_sample=False}) with no few-shot examples.

\begin{table}[h]
\centering
\caption{Exact prompt template for structured information extraction.}
\label{tab:prompt}
\small
\begin{tabular}{@{}p{0.92\linewidth}@{}}
\toprule
\texttt{You are a financial analyst. Extract structured}\\
\texttt{information from the following news article.}\\[4pt]
\texttt{Article: \{article\}}\\[4pt]
\texttt{Return a JSON object with exactly these fields:}\\
\texttt{\{\ "sentiment": "positive/negative/neutral",}\\
\texttt{\ \ "sentiment\_score": <float between -1 and 1>,}\\
\texttt{\ \ "event\_type": "earnings/merger/policy/}\\
\texttt{\ \ \ \ \ \ \ \ \ \ \ \ \ product/management/macro/other",}\\
\texttt{\ \ "impact\_subject": "company/industry/macro",}\\
\texttt{\ \ "time\_horizon": "short/long",}\\
\texttt{\ \ "confidence": <float between 0 and 1>\}}\\[4pt]
\texttt{Return only the JSON object, no explanation.}\\
\bottomrule
\end{tabular}
\end{table}

\end{document}